\documentclass[10pt,twocolumn,letterpaper]{article}

\usepackage[pagenumbers]{iccv} % To force page numbers, e.g. for an arXiv version

\usepackage{multirow}

\usepackage{tcolorbox}
\usepackage{tikz}

\usepackage{stfloats} % 添加到导言区

\definecolor{iccvblue}{rgb}{0.21,0.49,0.74}
\usepackage[pagebackref,breaklinks,colorlinks,allcolors=iccvblue]{hyperref}

\def\paperID{5221} % *** Enter the Paper ID here
\def\confName{ICCV}
\def\confYear{2025}

\title{Instruction-Oriented Preference Alignment for Enhancing Multi-Modal Comprehension Capability of MLLMs }

\author{\stepcounter{footnote}\normalsize{Zitian~Wang$^{1}$}
\qquad \normalsize{Yue~Liao$^{2}$}\thanks{Corresponding Author.} \qquad \normalsize{Kang~Rong} \qquad \normalsize{Fengyun~Rao} \qquad \normalsize{Yibo~Yang$^{3\dagger}$} \qquad \normalsize{Si~Liu$^{1}$} \\
    \small{$^{1}$Beihang University} \quad
	\small{$^{2}$National University of Singapore} \\
	\small{$^{3}$King Abdullah University of Science and Technology} \\
	{\small\tt $\{$wangzt.kghl,liaoyue.ai,yibo.yang93$\}$@gmail.com} \ \ {\small\tt $\{$liusi$\}$@buaa.edu.cn}
}

\begin{document}
\maketitle

\begin{abstract}

Preference alignment has emerged as an effective strategy to enhance the performance of Multimodal Large Language Models (MLLMs) following supervised fine-tuning. While existing preference alignment methods predominantly target hallucination factors, they overlook the factors essential for multi-modal comprehension capabilities, often narrowing their improvements on hallucination mitigation. To bridge this gap, we propose Instruction-oriented Preference Alignment (IPA), a scalable framework designed to automatically construct alignment preferences grounded in instruction fulfillment efficacy. Our method involves an automated preference construction coupled with a dedicated verification process that identifies instruction-oriented factors, avoiding significant variability in response representations. Additionally, IPA incorporates a progressive preference collection pipeline, further recalling challenging samples through model self-evolution and reference-guided refinement. Experiments conducted on Qwen2VL-7B demonstrate IPA's effectiveness across multiple benchmarks, including hallucination evaluation, visual question answering, and text understanding tasks, highlighting its capability to enhance general comprehension. The dataset is publicly available at
\url{https://huggingface.co/datasets/wangzt-kghl/IPA}.

\end{abstract}

\begin{figure*}[t!]
	\centering
	% \vspace{-4mm}
	\includegraphics[width=1\textwidth]{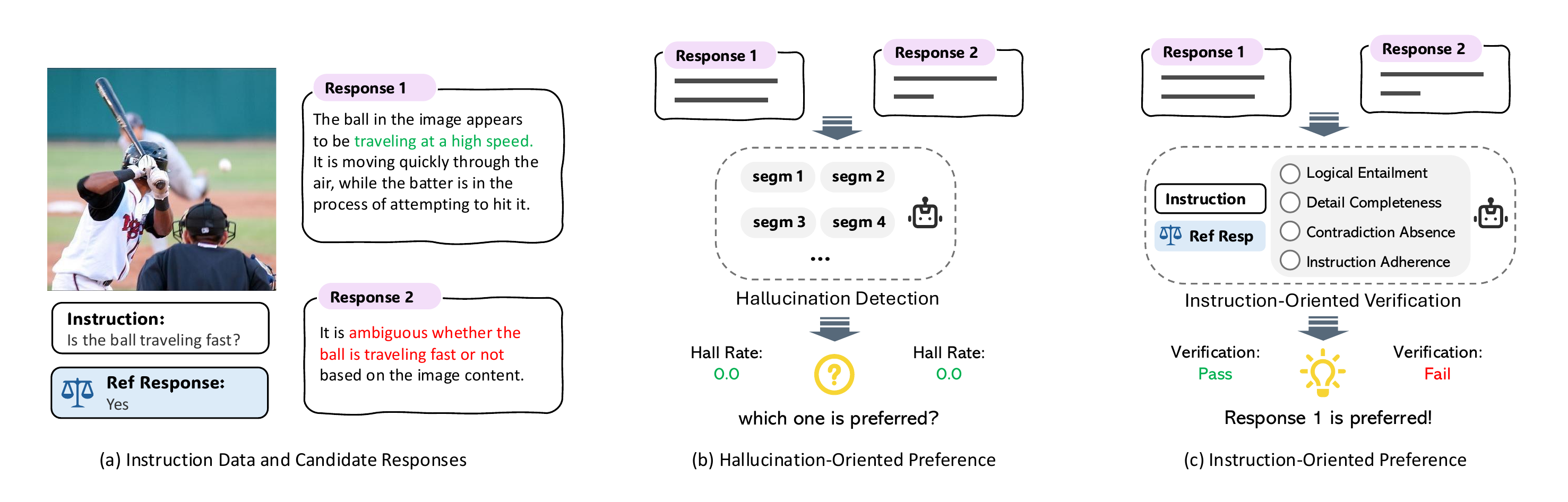}
	\vspace{-5mm}
	\caption{Different approaches for preference construction. (a) The instruction data and candidate responses. (b) Hallucination-oriented approaches construct preference pairs concerning the hallucination patterns, struggling to distinguish responses through intrinsic qualities.
    (c) Our approach introduces an instruction-oriented verification process for selecting preferred and dispreferred responses. The resulting preferences lead to enhanced comprehension capability driven by diverse instruction data. }
	% \vspace{-4mm}
	\label{fig:figure_1}
\end{figure*}

\section{Introduction}
\label{sec:intro}

Recent Multimodal Large Language Models~(MLLMs) have achieved impressive vision-language comprehension capabilities by scaling training data and model parameters~\cite{bai2025qwen25vltechnicalreport,liu2023visualinstructiontuning,team2023gemini}.  Supervised fine-tuning serves as the foundational mechanism for aligning vision and language into a unified space and enhancing the capabilities of MLLMs. However, noisy or ambiguous annotations in SFT data can cause models to learn incorrect information during training~\cite{sun2023aligninglargemultimodalmodels,yuksekgonul2023visionlanguagemodelsbehavelike,yu2024hallucidoctormitigatinghallucinatorytoxicity}.  
To mitigate this issue, preference alignment has emerged as a promising post-training strategy. 

The alignment signals within preference pairs reside in the factors that most differentiate the \emph{preferred} responses from the \emph{dispreferred} ones~\cite{ouyang2022training,yu2024rlhfvtrustworthymllmsbehavior}.
Models can then be aligned towards intended behaviors by learning these distinctions. Consequently, the central challenge lies in effectively identifying such discriminative factors to extract meaningful alignment signals.
A straightforward approach involves leveraging expert knowledge for annotation.
Common strategies primarily encompass human preference labeling~\cite{sun2023aligninglargemultimodalmodels,yu2024rlhfvtrustworthymllmsbehavior} and commercial model-based labeling (\emph{e.g.}, GPT-4V)~\cite{li2024vlfeedback}. 
However, both methods face scalability constraints: human annotation requires intensive labor costs, while reliance on commercial models incurs substantial financial expenses.
Recent research efforts have shifted towards scalable and automated preference construction.
Emerging methods either employ hallucination injection mechanisms to naturally induce preference pairs~\cite{zhou2024aligning,deng2024efficient,wang2024enhancingreasoningabilitymultimodal}, or detect hallucination patterns in candidate responses for preference construction~\cite{yu2024rlaif,ouali2024clip,liu2024mia,he2024topic}. 
While these approaches enable efficient model alignment at scale, they predominantly focus on the hallucination factors.

As depicted in Figure~\ref{fig:figure_1} (b), current hallucination-oriented methods neglect intrinsic quality dimensions and demonstrate limited discriminative power when differentiating responses, confining their improvements primarily to hallucination mitigation. This observation motivates our central research question:
\emph{What factors within the preference pairs critically determine the alignment towards enhanced general comprehension capabilities?} 
To address this, we propose identifying factors fundamentally associated with a response's efficacy in fulfilling instructions.
As shown in Figure~\ref{fig:figure_1} (c), we formulate the identification of these factors as an instruction-oriented verification process, establishing a clear decision boundary for distinguishing preferred versus dispreferred responses.
Driven by diverse instruction data, the resulting preferences lead to general improvements as the alignment is grounded in the MLLM's capability of comprehensive instruction fulfillment.

Our framework, Instruction-oriented Preference Alignment (IPA), comprises an automated preference construction mechanism and a progressive preference data collection pipeline. The automated preference construction operates through three sequential stages.
At the \emph{Sampling Stage}, we utilize generator models to sample multiple responses for given instructions.
At the \emph{Revision Stage}, we leverage the reviser models to identify critical flaws in the sampled responses and make appropriate refinements, driving the responses towards higher quality.
The \emph{Verification Stage} aims to quantify the instruction-oriented factors within candidate responses. 
A challenge arises from the inherent variability in response content and structure, which complicates the quantification process. 
We decouple this quantification from formatting constraints by harnessing the intrinsic evaluation capabilities of MLLMs, transforming it into a verification process.

Building upon the automated preference construction mechanism, we implement a progressive preference data collection pipeline.
In \emph{Round 1}, we apply the preference construction on a seed dataset and obtain a set of preference data. After initial preference construction, we optimize both generator and reviser models using collected preferences for capability enhancement.
In \emph{Round 2}, we re-process samples filtered as challenging from the first round using updated models, collecting additional preference data through self-evolution.
For the remaining hard samples after Round 2, \emph{Round 3} incorporates reference-guided refinement during revision, leveraging external knowledge to retrieve otherwise unattainable preference pairs.

We conduct comprehensive experiments on Qwen2VL-7B~\cite{yang2024qwen2} across various benchmark categories, including hallucination mitigation, general visual question answering, and text understanding.
Experimental results demonstrate consistent performance improvements over the baseline, indicating our method's potential for enhancing the general comprehension capabilities of MLLMs.

The contributions of this work are as follows: (1)~A scalable preference collection framework that transforms existing instruction data into alignment signals, steering alignment towards improved general comprehension capabilities driven by diverse instruction data. (2)~An instruction-oriented verification paradigm that identifies core underlying factors through clear decision boundaries, enabling preference construction on diverse data sources. (3)~Empirical validation across benchmarks spanning hallucination evaluation, general VQA, and text understanding, where consistent improvements indicate the feasibility and effectiveness of our method.

%-------------------------------------------------------------------------

\section{Related Work}

\noindent\textbf{Multimodal Instruction Data.}
Large-scale, high-quality training data is a cornerstone for building a advancing MLLM. However, it presents significant challenges to generate such data in resource constraints. Existing open-source large multimodal models including BLIP-2~\cite{li2023blip2bootstrappinglanguageimagepretraining}, Kosmos-2~\cite{peng2023kosmos2groundingmultimodallarge}, LLaVA~\cite{liu2024improvedbaselinesvisualinstruction,liu2023visualinstructiontuning}, Aquila-VL~\cite{gu2025infinitymmscalingmultimodalperformance}, MAmmoTH-VL~\cite{guo2024mammothvlelicitingmultimodalreasoning} typically use open source datasets in conjunction with internal data. There are three main approaches to processing these datasets. The first approach employs manual annotation, yielding precise and context-rich labels but proving prohibitively expensive and labor-intensive~\cite{xu2024visionflanscalinghumanlabeledtasks,deitke2024molmopixmoopenweights,mckinzie2024mm1methodsanalysis}, thus hindering scalability. The second approach leverages open-source datasets that are subsequently filtered and re-annotated by other open-source models~\cite{li2025eagle2buildingposttraining,chen2025expandingperformanceboundariesopensource}, enabling low-cost, large-scale expansion and iterative refinement through instruction templates and model refinement. However, this strategy often suffers from limited diversity and may introduce additional noise and hallucinations. The third approach involves proprietary closed-source models (e.g., GPT-4) to generate high-quality, diverse datasets but incurs substantial computational overhead and raises data privacy concerns~\cite{chen2024allava,chen2024sharegpt4v}. In this work, we propose a straightforward, scalable method that addresses these limitations by harnessing smaller open-source models, supplemented with instruction enhancement and preference learning, to cost-effectively construct high-quality multimodal datasets.

\begin{figure*}[t!]
	\centering
	% \vspace{-4mm}
	\includegraphics[width=1.0\textwidth]{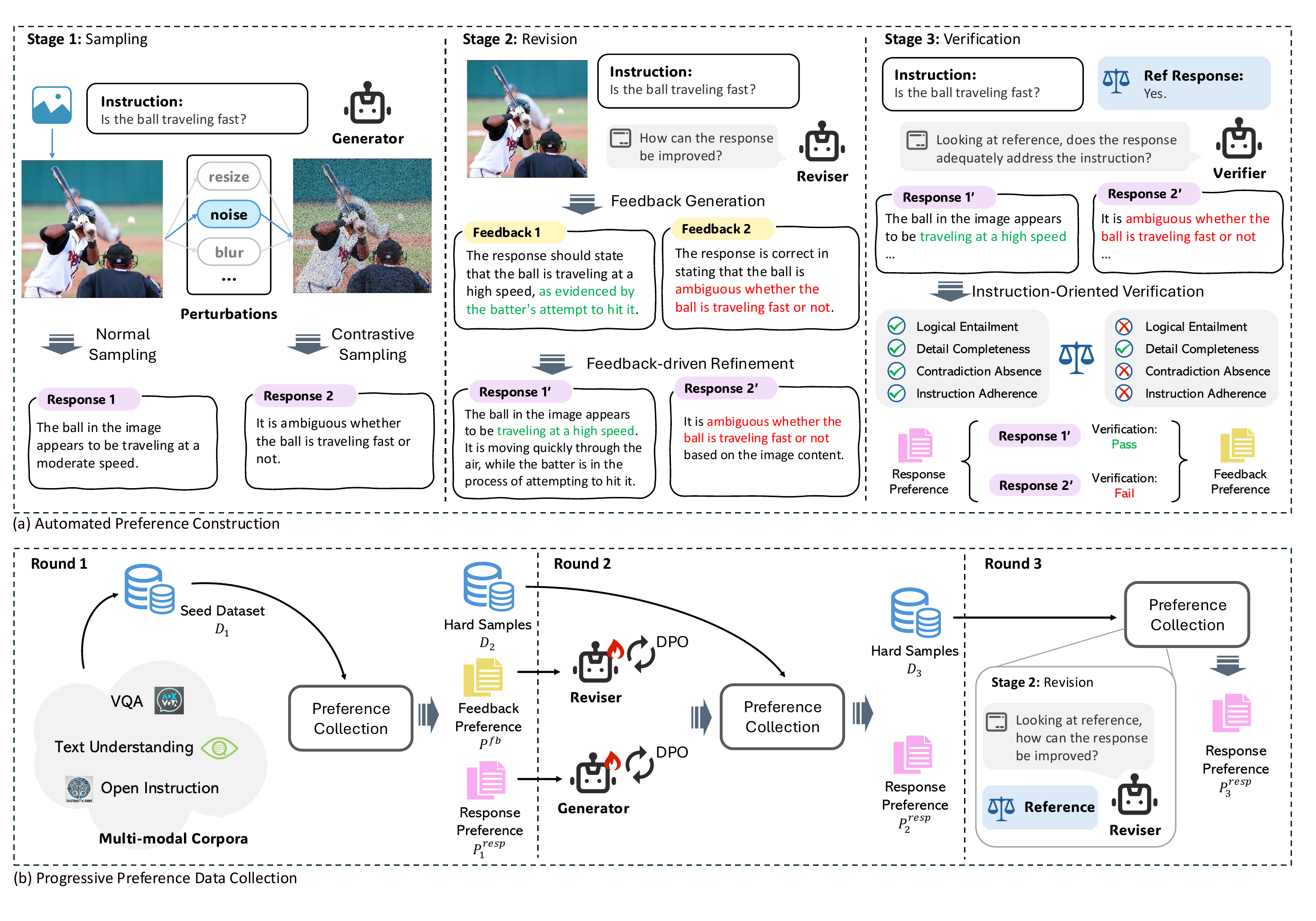}
	\vspace{-6mm}
	\caption{The framework of our approach. (a) The \textbf{automated preference construction} mechanism. The sampling stage generates diverse response candidates for given instructions with a contrastive sampling strategy. The revision stage enhances responses while expanding the depth of task-relevant information. The verification stage evaluates whether responses fulfill core instruction requirements, transforming alignment signal extraction into a verification task. (b) The \textbf{progressive preference data collection} pipeline. We implement a multi-round preference collection paradigm to maximize hard sample utilization with self-improvement and reference-guided improvement.}
	% \vspace{-4mm}
	\label{fig:framework}
\end{figure*}

\noindent\textbf{Preference Optimization.}
Despite remarkable progress, MLLMs still suffer from generating incorrect outputs misaligned with image context. Recent efforts have introduced preference optimization (PO) methods to mitigate this issue and enhance the output quality of both LLMs and MLLMs. InstructGPT~\cite{ouyang2022training} utilizes PPO algorithm~\cite{schulman2017proximal} guided by human feedback to produce more accurate and useful outputs. DPO~\cite{rafailov2023direct} employs an efficient Bradley-Terry-based~\cite{bradley1952rank} approach, removing the need for explicit reward models and significantly reducing computational overhead. LLaVA-RLHF~\cite{sun2023aligninglargemultimodalmodels} extends PO to the multimodal domain, utilizing human-ranked preferences to enhance visual dialogue abilities.  Furthermore, recent studies leverage reinforcement learning from human feedback~\cite{sun2023aligninglargemultimodalmodels,yu2024rlhfvtrustworthymllmsbehavior,gunjal2024detectingpreventinghallucinationslarge,yu2024rlhfvtrustworthymllmsbehavior} or AI feedback~\cite{li2023silkiepreferencedistillationlarge,zhao2024hallucinationsenhancinglvlmshallucinationaware,xiong2024llavacriticlearningevaluatemultimodal,wang2024enhancingreasoningabilitymultimodal,zhou2024aligning,deng2024efficient,yu2024rlaif,ouali2024clip,liu2024mia,he2024topic} to reduce hallucinations and strengthen vision language model (VLM) capabilities. These works primarily operate by constructing a data pipeline to generate batches of preference pairs, and subsequently PO to maximize the model's preference for positive samples.

\noindent\textbf{Large Models as Judge.} 
Constructing preference pairs for reinforcement learning, accuracy, and consistency are crucial. However, human annotations are often subjective, costly, and unstable. These limitations have motivated the emergence of the ``LLM-as-a-judge''~\cite{zheng2023judgingllmasajudgemtbenchchatbot,lin2023llm} approach, where large language models are employed to evaluate model generated outputs~\cite{zheng2023judgingllmasajudgemtbenchchatbot,zhu2025judgelmfinetunedlargelanguage}.
The scalability and adaptability LLMs make them particularly suitable for complex and large-scale evaluation tasks. Meanwhile, advancements in training paradigms such as reinforcement learning with human feedback (RLHF~\cite{ouyang2022training}) have enabled LLMs to become increasingly aligned with human values and reasoning processes. This alignment allows LLMs to assume more diverse roles, and they are now widely employed as Graders~\cite{dong2023raftrewardrankedfinetuning,wang2024halujcritiquebasedhallucinationjudge}, Critics~\cite{ke2023critiquellm,putta2024agent,xiong2024llava}, Verifiers~\cite{ling2023deductiveverificationchainofthoughtreasoning,shinn2023reflexion,wang2024speculative}, and reward/ranking models~\cite{luo2025wizardmathempoweringmathematicalreasoning,sun2023aligninglargemultimodalmodels,yang2024qwen2,yuan2023rrhfrankresponsesalign}.

\section{Method}

As depicted in Figure~\ref{fig:framework}, our framework IPA comprises an automated preference construction mechanism and a progressive preference collection pipeline. 

The automated preference construction mechanism comprises three stages. 
First, the sampling stage generates diverse response candidates for given instructions, with a contrastive sampling strategy that intentionally introduces quality discrepancies.
Second, the revision stage utilizes the reflection capabilities of MLLMs to improve responses by calibrating correctness and enriching the depth of task-relevant information.
Finally, the verification stage assesses whether responses satisfy core instruction requirements, thus converting alignment signal extraction into a verification task decoupled from response variations. 

Based on the preference construction mechanism, we propose progressively collecting preference data from existing multi-modal corpora, facilitating alignment in an effective and scalable manner.

\subsection{Sampling Responses for Instruction}  
\label{sec:sampling_stage}
We start from a dataset $\mathcal{D} = \{s_k \}_{k=1}^N$ created from open-source multi-modal corpora, where each sample $s = (V, I, r^{ref})$ contains an image $V$, an instruction $I$, and a reference response $r^{ref}$. For the given sample $ s \in \mathcal{D}$, we generate response candidates through two strategies:

\noindent\textbf{Normal Sampling.} The generator model $\pi_G$ (MLLMs) sample responses by:  
\begin{equation}
\label{equ:sampling}
    r^{norm} \sim \pi_G(\cdot \mid V, I; \theta_{\pi_G}),
\end{equation}  
yielding a response set $\mathcal{R}^{norm} = \{r^{norm}_i\}^{M}_{i=1}$ within the original output distribution, where $M$ denotes the number of sampling times.

\noindent\textbf{Contrastive Sampling.} 
Inspired by Visual Contrastive Decoding (VCD) methods~\cite{leng2023mitigatingobjecthallucinationslarge}, we utilize a contrastive sampling strategy to collect candidate responses with quality discrepancies.
Specifically, for the input image $V$, a perturbation operator $t $ is randomly selected from a predefined perturbation set $\mathcal{T}$ including noise, blur, resize, etc. Then the sampling is conditioned on the perturbed image:  
\begin{equation}
\label{equ:contrastive_sampling}
r^{{cont}} \sim \pi_G\big(\cdot \mid t(V), I; \theta_{\pi_G}\big), \quad \text{where } t \sim \mathcal{T},
\end{equation}
forming a contrastive response set $\mathcal{R}^{cont} = \{{r}^{cont}\}$.
The contrastive sampling strategy aims to enhance the diversity of flawed response patterns associated with $\pi_G$'s robustness in multi-modal comprehension, which can affect the ability of responses to maintain instruction fulfillment across different dimensions.

Note that we do not categorize $\mathcal{R}^{cont}$ as dispreferred responses.
Instead, it is combined with $\mathcal{R}^{norm}$ to form the initial response set  $\mathcal{R}^{init}= \mathcal{R}^{norm} \cup \mathcal{R}^{cont}$.

\subsection{Reflective Response Revision}
\label{sec:revision_stage}

The revision stage improves responses through two sequential phases: diagnostic feedback generation and feedback-driven refinement.

Given an initial response $r \in \mathcal{R}^{{init}}$ generated for a multimodal input $(V, I)$, the process first produces diagnostic feedback $fb$ through the reviser model $\pi_R$:
\begin{equation}
\label{equ:feedback_generation}
fb \sim \pi_R(\cdot \mid V, I, r, I^{fb}; \theta_{\pi_R}),
\end{equation}
where the reviser model is prompted by $I^{fb}$ to complement the initial responses.
The feedback not only identifies vision-related errors (i.e., visual hallucinations), but more crucially, expands the depth of task-relevant information within natural response distributions.
For instance, when addressing the question \textit{``Is there a cat in the image?''}, compared with the simple response \textit{``Yes.''}, an informative and evidentially grounded response \textit{``The image's lower-left corner displays a gray cat.''} more satisfies core instruction requirements by demonstrating observational logic and maintaining conversational fluency.

Then the reviser model subsequently refines the initial response through:  
\begin{equation}
r^{rev} \sim \pi_R(\cdot \mid V, I, r, fb, I^{rf}; \theta_{\pi_R}),
\end{equation}
where $I^{rf}$ is the prompt that directs the model to refine the initial response $r$ according to the information from $fb$.

This stage intends to enhance instruction-critical factors within the responses while preserving their natural output characteristics.  
We combine the initial response set $\mathcal{R}^{init}$ and revised response set $\mathcal{R}^{rev}$ to form a whole candidate response set $\mathcal{R} = \mathcal{R}^{rev} \cup \mathcal{R}^{init}$.

\subsection{Instruction-Oriented Response Verification}
\label{sec:verification_stage}
Given a sample $s=(V, I, r^{ref})$, our goal is to extract alignment signals and construct preference pairs from the candidate response set $\mathcal{R}$. 
To steer models towards enhanced general comprehension capabilities, we ground the preference assessment in the instruction fulfillment capacity: \textit{Does the response adequately address the instruction?} 

However, the inherent linguistic flexibility of natural language introduces confounding factors, where variations (e.g., stylistic differences, non-essential visual details, formatting conventions) obscure the differentiation of the truly preferred factors. 
To resolve the ambiguities, we establish a decision boundary for selecting preferred and dispreferred responses via an instruction-oriented verification process.

We operationalize this by leveraging the reference response $r^{ref}$ and the intrinsic evaluation capacity of large models~\cite{zheng2023judgingllmasajudgemtbenchchatbot,lin2023llm}. For each candidate response $r \in \mathcal{R}$, the verifier model $\pi_V$ is prompted to assess four different components:  
\begin{itemize}
    \item \emph{Logical Entailment:} Whether the reference answer $r^{ref}$ can be logically inferred from $r$ (i.e., $r \Rightarrow r^{ref}$).
    \item \emph{Detail Completeness:} Presence of missing critical details in $r$ compared to $r^{ref}$.
    \item \emph{Contradiction Absence:} Existence of contradictions between $r$ and $r^{ref}$.
    \item \emph{Instruction Adherence:} Adherence to the instruction $I$.
\end{itemize}

The verification process is then formalized as:  
\begin{equation}
v \sim \pi_V(\cdot \mid V, I, r, r^{ref}, I^{ver}; \theta_{\pi_V}),
\end{equation}
where $v \in \{0,1\}$ is a binary indicator, with $v=1$ signifying satisfaction of all conditions, $v=0$ indicating certain failure modes.

We partition $\mathcal{R}$ into preferred responses $\mathcal{R}^{win} = \{r \mid v=1\}$ and dispreferred responses $\mathcal{R}^{loss} = \{r \mid v=0\}$.  
The final preference pairs $\mathcal{P}^{resp} = \{(r^w, r^l)\}$ are constructed by pairing $r^w \in \mathcal{R}^{win}$ with $r^l \in \mathcal{R}^{loss}$ that explicitly differ in instruction fulfillment capacity, anchoring alignment signals to verifiable evidence.

\subsection{Progressive Preference Data Collection}  

We implement a progressive preference data collection pipeline for effective and scalable preference alignment.
This begins with constructing a seed dataset $\mathcal{D}$ comprising multi-source, multi-modal samples $(V, I, r^{ref})$. 
For datasets originally in multiple-choice format, we randomly convert some samples into open-ended questions by removing provided options from the instructions to enable more general comprehension. 
Consequently, $\mathcal{D}$ aggregates the following data sources:

\begin{itemize}
    \item \emph{Visual Question Answering (VQA)}: VQAv2~\cite{goyal2017makingvvqamatter}, OK-VQA~\cite{marino2019okvqavisualquestionanswering}, GQA~\cite{hudson2019gqanewdatasetrealworld}, A-OKVQA~\cite{schwenk2022aokvqabenchmarkvisualquestion}, M3CoT~\cite{chen2024m3cotnovelbenchmarkmultidomain}, InfoVQA~\cite{mathew2022infographicvqa}, ScienceQA~\cite{lu2022learn}, CLEVR-Math~\cite{lindström2022clevrmathdatasetcompositionallanguage}, ChartQA~\cite{masry2022chartqabenchmarkquestionanswering}, DocVQA~\cite{mathew2021docvqadatasetvqadocument}, DVQA~\cite{kafle2018dvqa}
    \item \emph{Text Understanding}: TextVQA~\cite{singh2019vqamodelsread}, LLaVAR~\cite{zhang2024llavarenhancedvisualinstruction}, TextCaps~\cite{sidorov2020textcapsdatasetimagecaptioning}  
    \item \emph{Open Instruction}: Infinity-MM~\cite{gu2025infinitymmscalingmultimodalperformance}
\end{itemize}

Since constructing valid preference pairs for all samples in $\mathcal{D}$ is infeasible (e.g., easy instructions may lack $\mathcal{R}^{loss}$ while hard ones lack $\mathcal{R}^{win}$), we employ a multi-round procedure to maximize hard sample utilization.

\noindent\textbf{Round 1.}  
Let the original dataset be $\mathcal{D}_1$. For each sample $s \in \mathcal{D}_1$, we construct preference pairs through Sections~\ref{sec:sampling_stage} to \ref{sec:verification_stage}, obtaining $\mathcal{R}^{win}$ and $\mathcal{R}^{loss}$. 
Samples failing to produce $\mathcal{R}^{win}$ (indicating higher difficulty) are retained to form $\mathcal{D}_2 = \{s \mid \mathcal{R}^{win}(s) = \emptyset, \forall s \in \mathcal{D}_1\}$.

\noindent\textbf{Round 2.}  
We adopt a self-improvement paradigm by optimizing the generator $\pi_G$ and reviser $\pi_R$ and then repeat the preference construction. 
Let $\mathcal{P}^{resp}_1 = \{(r^w, r^l)\}$ denote response preference pairs, and $\mathcal{P}^{fb}_1 = \{(fb^w, fb^l)\}$ represent feedback preference pairs obtained in Round 1. 
Specifically, $fb^w$ ($fb^l$) refers to feedback that leads to successful (failed) verification:
\begin{equation}
\begin{aligned}
fb^w \in \{ fb \mid r^{rev} \in \mathcal{R}^{win},  r^{rev} \sim \pi_R(\cdot \mid V, I, r, fb, I^{rf}) \}, \\
fb^l \in \{ fb \mid r^{rev} \in \mathcal{R}^{loss}, r^{rev} \sim \pi_R(\cdot \mid V, I, r, fb, I^{rf}) \}.
\end{aligned}
\end{equation}

The models are optimized via DPO:
\begin{equation}
\begin{aligned}
    \pi_G^+ = \text{DPO}(\pi_G, \mathcal{P}^{resp}_1), \quad
    \pi_R^+ = \text{DPO}(\pi_R, \mathcal{P}^{fb}_1).
\end{aligned}
\end{equation}
Then the enhanced models $\pi_G^+$ and $\pi_R^+$ reprocess $\mathcal{D}_2$ to collect second-round preference pairs $\mathcal{P}^{resp}_2$.

\begin{figure}[t!]
	\centering
	% \vspace{-5mm}
	\includegraphics[width=0.49\textwidth]{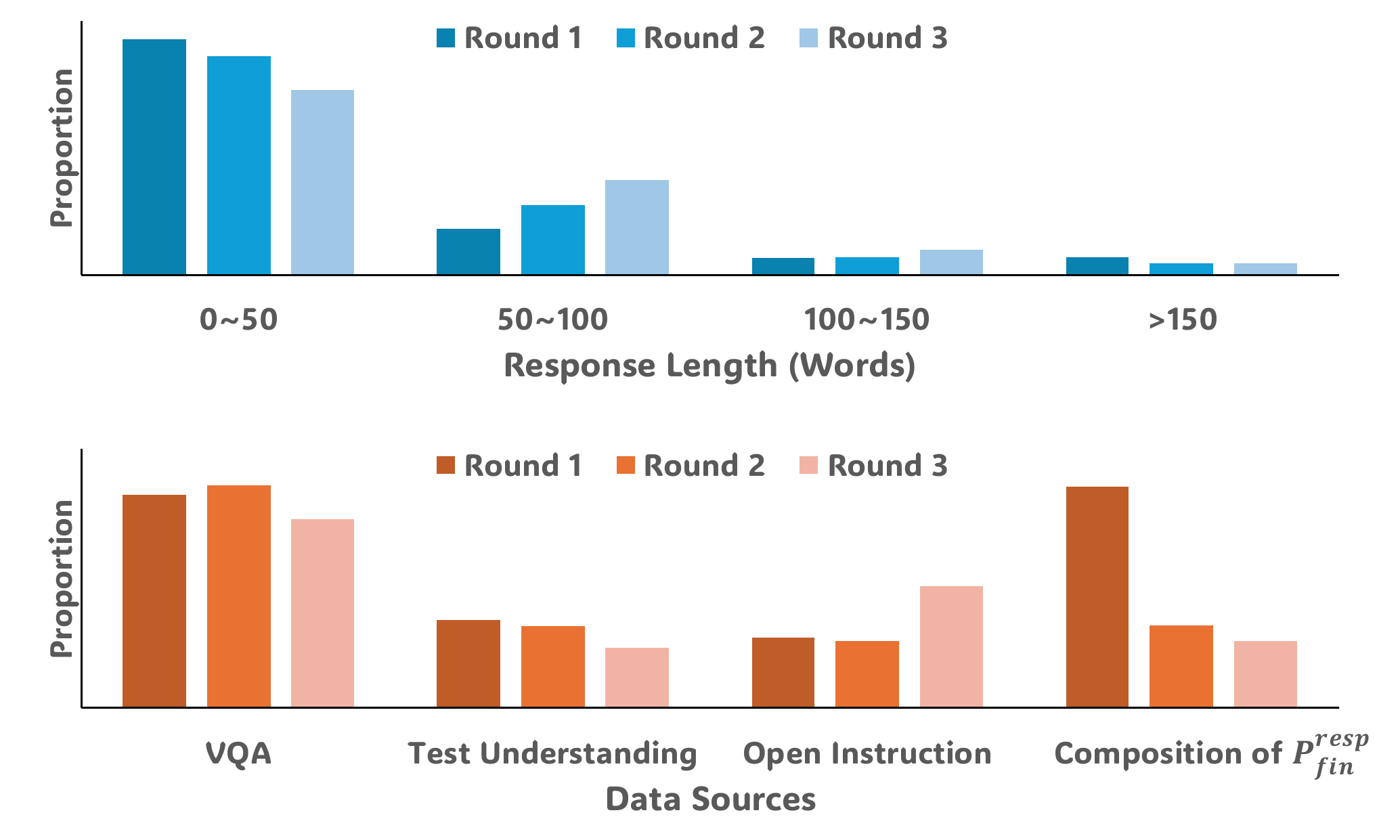}
	\vspace{-2mm}
	\caption{Preference data distribution in different rounds. The preference collection pipeline retrieves samples with longer responses, which indicate more complex scenarios.}
	\label{fig:stats}
	\vspace{0mm}
\end{figure}

\noindent\textbf{Round 3.} For the remaining challenging samples $\mathcal{D}_3$,  we augment the reviser with reference-guided feedback:  
\begin{equation}
fb \sim \pi_R(\cdot \mid V, I, r, r^{ref}, I^{fb}; \theta_{\pi_R}),
\end{equation}
where $r^{ref}$ serves as an anchor for critical feedback, in contrast to Equation~\eqref{equ:feedback_generation}. An alternative approach is to augment the generator with reference-guided sampling:
\begin{equation}
\label{equ:reference_guided_sampling}
r^{norm} \sim \pi_G(\cdot \mid V, I, r^{ref}; \theta_{\pi_G}),
\end{equation}
but we empirically observe that it sometimes induces simple imitation of $r^{ref}$ in the response $r^{norm}$.

The final preference data combines three-round results $\mathcal{P}^{resp}_{fin} = \bigcup_{t=1}^3 \mathcal{P}^{resp}_t$, which comprises 89k preference pairs.
Figure~\ref{fig:stats} illustrates the data statistics across different rounds. Increased average length of responses across different rounds indicates that our progressive pipeline can enhance the utilization of more challenging samples. 
We validate the design choices of this pipeline in Section~\ref{sec:ablation}.

\begin{table*}[t]
\begin{center}
\resizebox{1.0\textwidth}{!}{
\renewcommand{\arraystretch}{1.2}
\setlength{\tabcolsep}{1.0mm}

\begin{tabular}{l | c c c| c c c  c c |c}
\toprule[1.5pt]

\multirow{2}{*}{\textbf{Model}} & \multicolumn{3}{c|}{\textbf{Hallucination}} & \multicolumn{5}{c|}{\textbf{General VQA}} & \multicolumn{1}{c}{\textbf{Text}} \\ 
% \cmidrule(l){2-4} \cmidrule(l){5-6} \cmidrule(l){7-9}

 & HallBench$_{avg}$ & POPE & MMHal & MMMU$_{val}$ & MMStar & MMVet$_{turbo}$ & MME$_P$ & LLaVABench & OCRBench \\ 
\midrule
MiniCPM-V 2.6~\cite{yao2024minicpm}  & 48.1 & - & - & 49.8 & - & 60.0 & - & - & 85.2 \\ 
LLaVA-OV-7B~\cite{li2024llavaonevisioneasyvisualtask} & - & - & - & 48.8 & 61.7 & 57.5 & 1589.0 & 90.7 & - \\ 
InternVL2-8B~\cite{chen2024far} & 45.2 & - & - & 51.8 & 62.0 & 54.2 & - & - & 79.4 \\ 
InternVL2.5-8B~\cite{chen2025expandingperformanceboundariesopensource} & 50.1 & 90.6 & 3.7 & 56.0 & 62.8 & 62.8 & - & - & 82.2 \\ 
Qwen2VL-7B~\cite{yang2024qwen2} & 50.6 & - & - & 54.1 & 60.7 & 62.0 & - & - & 84.5 \\ 
Qwen2.5VL-7B~\cite{bai2025qwen2} & 52.9 & - & - & 58.6 & 63.9 & 67.1 & - & - & 86.4 \\ 
\midrule
Qwen2VL-7B$\dag$ & 50.0 & 86.2 & 3.6 & 53.8 & 60.7 & 63.1 & 1676.3 & 76.6 & 86.2 \\ 
~~ + IPA (ours) & 54.3 \textcolor{red}{\scriptsize{+4.3}} & 87.2 \textcolor{red}{\scriptsize{+1.0}} & 3.7 \textcolor{red}{\scriptsize{+0.1}} & 54.6 \textcolor{red}{\scriptsize{+0.8}} & 61.7 \textcolor{red}{\scriptsize{+1.0}} & 64.2 \textcolor{red}{\scriptsize{+1.1}} & 1687.4 \textcolor{red}{\scriptsize{+11.1}} & 84.0 \textcolor{red}{\scriptsize{+7.4}} & 87.3 \textcolor{red}{\scriptsize{+1.1}} \\ 
Qwen2.5VL-7B$\dag$ & 54.7 & 86.2 & 3.7 & 57.6 & 64.7 & 65.6 & 1694.4 & 75.2 & 87.5 \\ 
~~ + IPA (ours) & 55.7 \textcolor{red}{\scriptsize{+1.0}} & 86.6 \textcolor{red}{\scriptsize{+0.4}} & 3.9 \textcolor{red}{\scriptsize{+0.2}} & 59.8 \textcolor{red}{\scriptsize{+2.2}} & 66.5 \textcolor{red}{\scriptsize{+1.8}} & 68.3 \textcolor{red}{\scriptsize{+2.7}} & 1706.5 \textcolor{red}{\scriptsize{+12.1}} & 87.3 \textcolor{red}{\scriptsize{+12.1}} & 88.2 \textcolor{red}{\scriptsize{+0.7}} \\

\bottomrule[1.5pt]
\end{tabular}

}
\caption{
The evaluation results on different benchmarks.
The upper part are the results reported in the original papers, while the lower part are our reproduced results (with $\dag$). Our method brings consistent improvements across 9 benchmarks compared to the base models, spanning three different dimensions: hallucination evaluation, general VQA, and text understanding.  
}
\label{tab:table_1}
\end{center}
\vspace{-5mm}
\end{table*}

\section{Experiments}

\subsection{Implementation Details}

\noindent\textbf{Model Configurations.}
We adopt Qwen2VL-7B~\cite{yang2024qwen2} as the base model architecture for preference data collection and performance validation, serving as the backbone for the generator, reviser, and verifier models. Parameters of different models are not shared.
To enhance response diversity, we introduce additional models with comparable or weaker capabilities for response sampling (Equation~\ref{equ:sampling}-\ref{equ:contrastive_sampling}) and feedback generation (Equation~\ref{equ:feedback_generation}), including Qwen2VL-2B~\cite{yang2024qwen2}, InternVL2-2B$\&8$B~\cite{chen2024far}, and LLaVA-OV-0.5B$\&$7B~\cite{li2024llavaonevisioneasyvisualtask}.

\noindent\textbf{Training and Evaluation.}
For all experiments in this section, we adopt the same training protocol and hyper-parameters to ensure fair comparison.
The alignment process is implemented via widely adopted Direct Preference Optimization (DPO)~\cite{rafailov2023direct} with Low-Rank Adaptation (LoRA)~\cite{hu2022lora}. The models are both trained for 1 epoch. 
We evaluate model performance across most benchmarks using VLMEvalKit~\cite{duan2024vlmevalkit}, an open-sourced evaluation toolkit adopted by OpenCompass Leaderboard~\cite{2023opencompass}. For evaluation on MMHal-Bench~\cite{sun2023aligninglargemultimodalmodels}, we use the official codebase.

More details and qualitative analysis are presented in the supplementary material.

\subsection{Benchmarks}
We evaluate the models across three critical dimensions: general visual question answering (VQA), hallucination, and text understanding.

For general VQA, we evaluate on five benchmarks:
(1) \emph{MMStar}~\cite{chen2024rightwayevaluatinglarge}, which benchmarks core multi-modal capabilities of MLLMs with carefully balanced and purified samples.
(2) \emph{MMMU}~\cite{yue2024mmmumassivemultidisciplinemultimodal}, which tests multimodal understanding in multi-disciplinary contexts.
(3) \emph{MMVet}~\cite{yu2024mmvetevaluatinglargemultimodal}, which evaluates open-ended visual question answering, specializing in fine-grained and comprehensive visual understanding and compositional reasoning.
(4) \emph{MME}~\cite{fu2024mmecomprehensiveevaluationbenchmark}, which provides a comprehensive evaluation covering perceptual and cognitive capabilities.
(5) \emph{LLaVABench}~\cite{li2024llavanextinterleavetacklingmultiimagevideo}, which measures instruction-following capabilities in diverse visual contexts.

For hallucination, we evaluate on three benchmarks:
(1) \emph{HallusionBench}~\cite{guan2024hallusionbenchadvanceddiagnosticsuite}, which quantifies visual-language hallucination through controlled contrastive samples.
(2) \emph{POPE}~\cite{li2023evaluatingobjecthallucinationlarge}, which measures hallucination rates through object existence verification in controlled settings.
(3) \emph{MMHal-Bench}~\cite{sun2023aligninglargemultimodalmodels}, which focuses on multi-modal hallucination patterns, including contextual mismatch and spurious detail generation.

For text understanding, we evaluate on 
\emph{OCRBench}~\cite{Liu_2024}, a specialized benchmark for scene text recognition, document text recognition, and key information extraction.

\begin{table*}[t]
\begin{center}
\resizebox{1.0\textwidth}{!}{
\renewcommand{\arraystretch}{1.2}
\setlength{\tabcolsep}{1.0mm}

\begin{tabular}{l | c c c| c c c  c c |c}
\toprule[1.5pt]

\multirow{2}{*}{\textbf{Model}} & \multicolumn{3}{c|}{\textbf{Hallucination}} & \multicolumn{5}{c|}{\textbf{General VQA}} & \multicolumn{1}{c}{\textbf{Text}} \\ 
 & HallBench$_{avg}$ & POPE & MMHal & MMMU$_{val}$ & MMStar & MMVet$_{turbo}$ & MME$_P$ & LLaVABench & OCRBench \\ 
\midrule

Qwen2VL-7B & 50.0 & 86.2 & 3.6 & 53.8 & 60.7 & 63.1 & 1676.3 & 76.6 & 86.2 \\ 
~~ + RLHF-V & 50.2 & 86.4 & 3.5 & 52.9 & 61.0 & 60.4 & 1682.0 & 76.6 & 86.3 \\ 
~~ + RLAIF-V & 50.2 & 86.5 & 3.4 & 53.8 & 61.1 & 58.2 & 1674.1 & 78.8 & 86.4 \\ 
~~ + VLFeedback & 52.2 & 84.7 & 3.6 & 53.7 & 60.7 & 60.6 & 1682.4 & 81.4 & 86.6 \\ 
~~ + Topic-Overwrite & 50.3 & 86.4 & 3.5 &53.6 & 60.9 & 60.0 & 1682.0 & 78.9 & 86.1 \\ 
~~ + MMPR & 53.4 & 86.4 & \textbf{3.9} & 54.3 & 58.5& 61.7 & 1680.5 & 83.5 & 86.3 \\ 
~~ + IPA (ours) & \textbf{54.3} & \textbf{87.2} & 3.7 & \textbf{54.6} & \textbf{61.7} & \textbf{64.2} & \textbf{1687.4} & \textbf{84.0} & \textbf{87.3} \\

\bottomrule[1.5pt]
\end{tabular}

}
\caption{
Comparison with existing preference alignment methods on Qwen2VL-7B. All the results are produced under the same training and evaluation protocol.
}
\label{tab:table_qwen}
\end{center}
\vspace{-1mm}
\end{table*}

\begin{table*}[t]
\begin{center}
\resizebox{1.0\textwidth}{!}{
\renewcommand{\arraystretch}{1.3}
\setlength{\tabcolsep}{0.9mm}

\begin{tabular}{l| c | c c c c c c  c }
\toprule[1.5pt]
 Model & Average & HallBench$_{avg}$ & MMMU$_{val}$ & MMStar & {MMVet}$_{turbo}$ &  OCRBench  & POPE & LLaVABench  \\
\midrule

Qwen2VL-7B & 68.1  & 50.0  & 53.8  & 60.7  & 63.1  & 86.2  & 86.2  & 76.6  \\
~~ + Round 1 w/o CS & 69.1  & 52.3  & 53.4  & 61.2  & 61.7  & 87.0  & 86.7  & 81.3  \\  
~~ + Round 1 & 70.1  & 52.9  & 54.2  & {61.9} & 62.7  & 87.0  & {87.2}  & 84.6  \\ 
~~ + Round 1 + Round 2 & 70.4  & 53.7  & 54.6  & {61.7}  & 63.8  & 86.9  & 87.1  & {85.2}  \\ 
~~ + Round 1 + Round 2 + Round 3 w/ RGS & 69.8  & 53.4  & {54.7}  & 61.1  & 62.7  & 87.2  & 86.6  & 83.1  \\ 
~~ + Round 1 + Round 2 + Round 3  & \textbf{70.5}  & {54.3}  & 54.6  & {61.7}  & {64.2}  & {87.3}  & {87.2}  & 84.0  \\

\bottomrule[1.5pt]
\end{tabular}

}
\caption{
Ablation study on the design choices of our approach. \emph{CS} denotes the contrastive sampling (Equation~\ref{equ:contrastive_sampling}). \emph{RGS} denotes the reference-guided sampling (Equation~\ref{equ:reference_guided_sampling}). 
}
\label{tab:table_abl}
\end{center}
\vspace{-4mm}
\end{table*}

\subsection{Preference Alignment Methods}

We select five representative approaches for comparison: 
(1) \emph{RLHF-V}~\cite{yu2024rlhfvtrustworthymllmsbehavior}, which targets improving the trustworthiness of MLLMs through fine-grained correctional human feedback.
(2) \emph{RLAIF-V}~\cite{yu2024rlaifvopensourceaifeedback}, which counts on the divide-and-conquer decomposition strategy for automated trustworthiness identification.
(3) \emph{VLFeedback}~\cite{li2024vlfeedbacklargescaleaifeedback}, which builds a diverse pool comprising 12 MLLMs for response generation, following by GPT-4V to annotate response quality concerning helpfulness, visual faithfulness, and ethics.
(4) \emph{Topic-Overwrite}~\cite{he2024topiclevelselfcorrectionalapproachmitigate}, which performs topic-level decomposition and overwrites problematic response patterns to mitigate hallucination.
(5) \emph{MMPR}~\cite{wang2024enhancingreasoningabilitymultimodal}, which collects three million preference pairs by formatted answer matching and dropout next-token prediction, mainly focusing on improving the Chain-of-Thought (CoT) reasoning performance of MLLMs.

\subsection{Main Results}
\label{sec:exp_main_results}

Our main experimental results are presented in Table~\ref{tab:table_1}, which involves comparison with existing state-of-the-art models. The upper part of the table reports evaluation results taken directly from the original papers, while the lower part presents our reproduced results with the same evaluation protocol. Although slight discrepancies exist between the numbers, our standardized evaluation protocol ensures fair comparisons.

We first perform preference alignment on Qwen2VL-7B via DPO, achieving consistent improvements across nine benchmarks.
Although our approach is not specifically designed for hallucination mitigation, it achieves a 4.3-point improvement on HallucinationBench, indicating its effectiveness in mitigating both linguistic hallucination and visual illusion. Additional improvements of 1.0 on POPE and 0.1 on MMHal-Bench further confirm our method's capability in reducing hallucination tendencies.
For general VQA benchmarks assessing diverse MLLM capabilities, our method demonstrates comprehensive improvements: 0.8 on MMMU, 1.0 on MMStar, 1.1 on MMVet, 11.1 on MME perception, and 7.4 on LLaVABench. These results substantiate our method's effectiveness in enhancing general visual comprehension abilities. For text understanding, we observe 1.1 point improvement on OCRBench, suggesting that appropriate expansion of data sources could further extend our method's benefits to additional domains.

To verify generality, we apply the collected preferences to align a more capable model, Qwen2.5VL-7B~\cite{bai2025qwen2}. The improvements on Qwen2.5VL-7B remain consistent across all benchmarks, demonstrating effective knowledge transfer. 
This suggests that the instruction-oriented factors identified by our method possess generalized applicability.

\subsection{Comparison with Existing Methods}

We conduct comprehensive comparisons with existing preference alignment approaches to validate the advantage of our method in enhancing general visual comprehension capabilities. 
The comparisons are utilize the same base model and training/evaluation protocol.

As shown in Table~\ref{tab:table_qwen}, our method achieves improvements on most benchmarks compared to existing methods.
For hallucination mitigation, our approach achieves the highest score on HallusionBench (54.3 vs. 53.4) and optimal POPE accuracy (87.2 vs. 86.5), while MMPR shows better MMHal performance. 
In general VQA benchmarks, we achieve superior performances on MMMU, MMStar, MMVet, MME perception, and LLaVABench, outperforming the baseline methods. 
For text understanding, our method also achieves a higher OCRBench score of 87.3.

Empirical analysis reveals a divergence in alignment philosophies among different methods. Methods specifically focused on particular aspects of capability, such as hallucination mitigation (e.g., RLHF-V, RLAIF-V, Topic-Overwrite), tend to exhibit performance trade-offs across multi-dimensional evaluation criteria.
In contrast, our method demonstrates consistent improvements across diverse benchmarks, notably surpassing VLFeedback and MMPR in general visual comprehension tasks.

These results suggest that our instruction-oriented paradigm can achieve a more holistic optimization when preference alignment is grounded in the intrinsic capability of MLLMs for comprehensive instruction fulfillment.

\subsection{Ablation Study}
\label{sec:ablation}

The results in Table~\ref{tab:table_abl} demonstrate the impact of each design choice in our approach through controlled experiments.

\noindent\textbf{Base Components Effectiveness.} 
The initial alignment round (Round 1) achieves 70.1 average score, providing 2.0 point absolute improvement over the baseline. Notably, the contrastive sampling strategy contributes 1.0 point gain on average, particularly enhancing performance on hallucination-sensitive benchmarks (+2.9 on HallusionBench) by exposing diverse error patterns through image perturbations.

\noindent\textbf{Self-Improvement Paradigm.} The second round of alignment (Round 1+2) yields an additional 0.3 point average improvement over the first round from 70.1 to 70.4. This validates our hypothesis that the self-improvement of both generator and reviser models enables better handling of more challenging samples, benefiting the production of preferred alignment signals.

\noindent\textbf{Reference-Guided Augmentation.}  
The full three-round pipeline (Round 1+2+3) achieves best performance, demonstrating the efficacy of reference-guided feedback generation in retrieving hard samples. 
Notably, naive reference-guided sampling degrades performance by 0.7.
This aligns with our observation that directly conditioning on reference $r^{ref}$ during response sampling sometimes induces the pattern imitation towards $r^{ref}$, which may compromise the diversity in response distribution.

\begin{table}[t]
\begin{center}
\resizebox{0.49\textwidth}{!}{
\renewcommand{\arraystretch}{1.2}
\setlength{\tabcolsep}{1.0mm}

\begin{tabular}{l| c c c c c }
\toprule[1.5pt]
 Model & HallBench & POPE & MMMU & MMStar &  MMVet   \\
\midrule

LLaVA-1.5-7B & 25.9 & 84.8 & 36.3 & 32.4 & 29.2 \\ 
~~ + RLHF-V & 25.9 & 84.6 & 36.7 & 32.6 & 29.6 \\ 
~~ + RLAIF-V & 24.6 & 85.4 & 36.8 & 32.9 & 28.7 \\ 
~~ + VLFeedback & \textbf{32.0} & 74.8 & 36.8 & \textbf{33.5} & 32.8 \\ 
~~ + Topic-Overwrite & 25.8 & 84.5 & 36.7 & 32.8 & 31.3 \\ 
~~ + IPA$_{sub}$ & 26.7 & \textbf{85.7} & \textbf{37.4} & 33.3 & \textbf{34.1} \\ 
        
\bottomrule[1.5pt]
\end{tabular}

}
\vspace{-0mm}
\caption{
Comparison with existing preference alignment methods on LLaVA-1.5-7B with the same training and evaluation protocol.
}
\label{tab:table_llava}
\end{center}
\vspace{-4mm}
\end{table}

\noindent\textbf{Accumulated Improvement.}  
The progressive pipeline unlocks 2.4 point overall improvement through three alignment rounds. All rounds contribute to an accumulated improvement: Round 1 establishes fundamental alignment, Round 2 involves self-evolution to produce higher-quality alignment signals, and Round 3 further recalls hard samples through reference-guided revision. 
These results validate the effectiveness of the progressive alignment pipeline. 

\subsection{Generality of Our Method}
While our preference data is primarily constructed with Qwen2VL-7B, we further validate its generalizability by applying it to align models with different capacity levels. 
Our evaluation encompasses both a more capable model Qwen2.5VL-7B (as detailed in Section~\ref{sec:exp_main_results}) and a less capable model LLaVA-1.5-7B~\cite{liu2024improvedbaselinesvisualinstruction}.

For LLaVA-1.5-7B, we observe performance degradation when directly applying the curated preference data for alignment, which we conjecture is due to its constrained visual perception capabilities.
Specifically, LLaVA-1.5-7B's weaker vision encoder (limited to 336$\times$336 resolution) may struggle to percept the detailed visual elements in the data.
Taking this into consideration, we simply filter out samples with response longer than 150 words, and create a subset (denoted as IPA$_{sub}$) for the experiments.

As shown in Table~\ref{tab:table_llava}, our approach achieves balanced overall improvements across capability dimensions. 
Compared to the base model, we observe moderate gains in hallucination mitigation (HallusionBench and POPE) and general visual comprehension (MMMU and MMVet).
Our method also avoids the severe accuracy-consistency trade-offs---while VLFeedback shows stronger HallusionBench and MMStar performance, it compromises POPE score (-10.0).
Given these experimental results, the successful knowledge transfer to stronger and weaker models demonstrates the effectiveness of our core design principle.

\section{Conclusion}
This work introduces Instruction-oriented Preference Alignment (IPA), a scalable preference alignment framework designed for MLLMs. IPA employs a progressive preference collection pipeline, leveraging instruction-oriented verification to efficiently extract high-quality alignment signals. Extensive evaluations across multiple benchmarks demonstrated IPA’s effectiveness, highlighting its potential for improving multimodal comprehensive capabilities. 
Due to computational constraints, exploring IPA’s application using more powerful models (e.g., with tens of billions of parameters) remains an avenue for future research.

{
    \small
    \bibliographystyle{ieeenat_fullname}
    \bibliography{main}
}

\appendix

\clearpage

\newpage

\begin{table*}[b]
\begin{center}

\caption{
Experiments with more advanced models. The training dataset is the same as in the main manuscript. 
% $\dag$ denotes the reproduced results.
}
\small
% \vspace{-1mm}
\resizebox{1.0\textwidth}{!}{
\renewcommand{\arraystretch}{1.0}
\setlength{\tabcolsep}{1.0mm}
\begin{tabular}{l | c c c| c c c  c c |c}
\toprule[1.5pt]

\multirow{2}{*}{\textbf{Model}} & \multicolumn{3}{c|}{\textbf{Hallucination}} & \multicolumn{5}{c|}{\textbf{General VQA}} & \multicolumn{1}{c}{\textbf{Text}} \\ 

 & HallBench$_{avg}$ & POPE & MMHal & MMMU$_{val}$ & MMStar & MMVet$_{turbo}$ & MME$_P$ & LLaVABench & OCRBench \\ 
\midrule

InternVL3-14B$\dag$ & 55.7 & 89.5 & 3.8 & 63.1 & 68.8 & 73.0 & 1740.5 & 83.0 & 87.6 \\ 
~~ + IPA (ours) &  56.0 \textcolor{red}{\scriptsize{+0.3}}& 89.6 \textcolor{red}{\scriptsize{+0.1}}& 3.9 \textcolor{red}{\scriptsize{+0.1}}& 63.3 \textcolor{red}{\scriptsize{+0.2}}& 69.1 \textcolor{red}{\scriptsize{+0.3}}& 75.3 \textcolor{red}{\scriptsize{+2.3}}& 1742.3 \textcolor{red}{\scriptsize{+1.8}}& 89.9 \textcolor{red}{\scriptsize{+6.9}}& 88.0\textcolor{red}{\scriptsize{+0.4}} \\ 
\midrule
Qwen2.5VL-72B$\dag$ & 55.3 & 86.4 & 4.3 & 68.1 & 71.6 & 74.4 & 1721.3 & 97.6 & 88.8 \\ 
~~ + IPA (ours) & 58.4 \textcolor{red}{\scriptsize{+3.1}} & 86.9 \textcolor{red}{\scriptsize{+0.5}} & 4.4 \textcolor{red}{\scriptsize{+0.1}} & 69.0 \textcolor{red}{\scriptsize{+0.9}} & 71.5 \textcolor{green}{\scriptsize{-0.1}} & 74.6 \textcolor{red}{\scriptsize{+0.2}} & 1730.8 \textcolor{red}{\scriptsize{+9.5}} & 102.2 \textcolor{red}{\scriptsize{+4.6}} & 89.2 \textcolor{red}{\scriptsize{+0.4}} \\
              
\bottomrule[1.5pt]
\end{tabular}
}
% \vspace{-2mm}
\label{tab:table_large_models}
\end{center}
% \vspace{-5mm}
\end{table*}

\section{More Implementation Details}

\subsection{Training Configuration}
For model optimization, we leverage Direct Preference Optimization (DPO) with Low-Rank Adaptation (LoRA).
Specifically, we use the AdamW optimizer with a cosine learning rate scheduler, starting from the initial learning rate of 5e-6. The total batch size is set to 8 and the gradient accumulation step is 8, resulting in an effective batch size of 64. 
All the models are trained for 1 epoch in the comparisons.

\subsection{Prompt Design}

We detail the prompts used in the automated preference construction.
Specifically, these comprises the prompt $I^{fb}$ for feedback generation (Equation 3), the prompt $I^{rev}$ for response refinement (Equation 4), and the prompt $I^{ver}$ for instruction-oriented verification (Equation 5).
The prompts are displayed in Table~\ref{tab:prompt_feedback}, \ref{tab:prompt_refinement} and \ref{tab:prompt_verification} respectively.

\section{Qualitative Analysis}

\begin{figure*}[t!]
	\centering
	\vspace{-3mm}
	\includegraphics[width=0.93\textwidth]{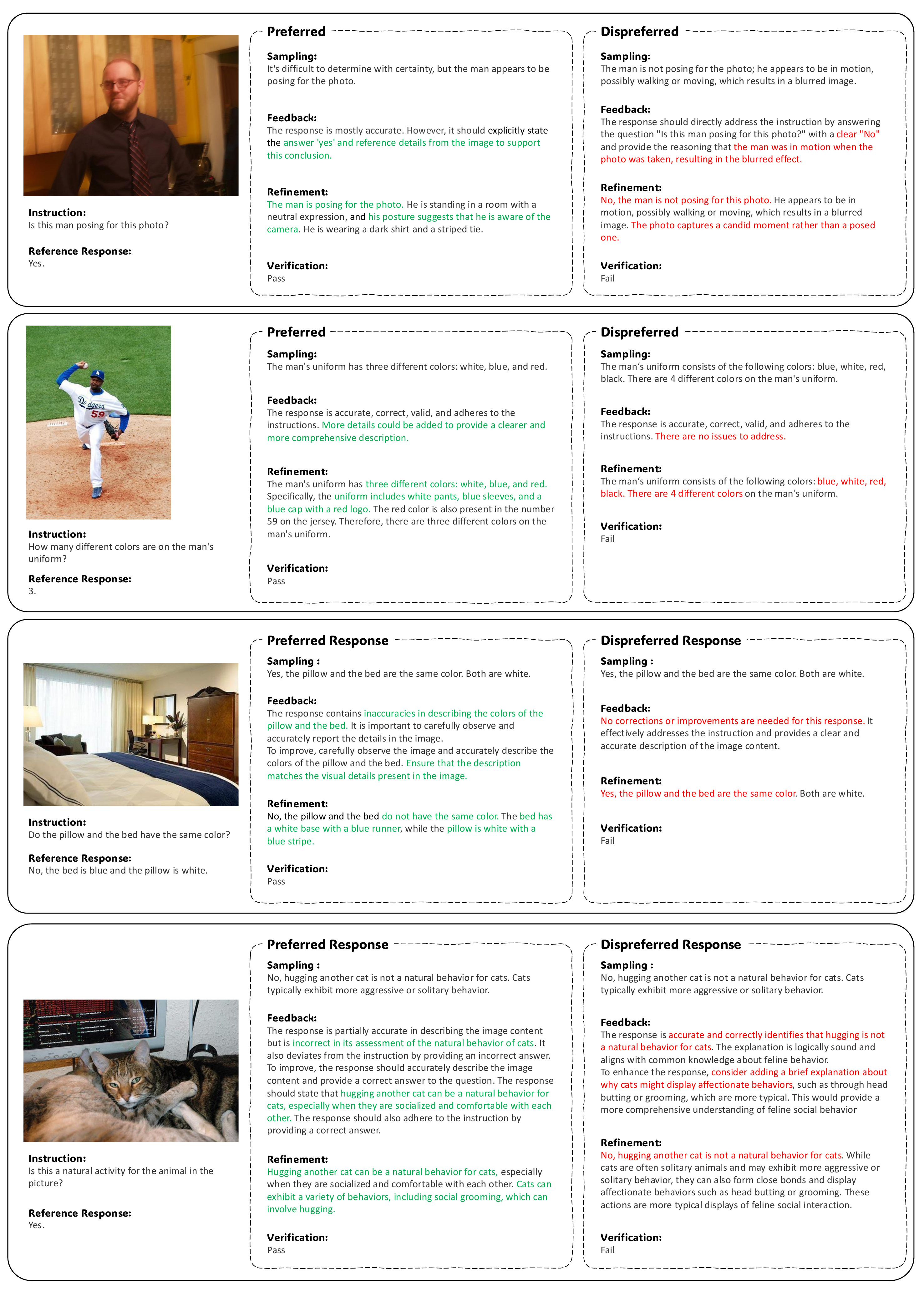}
	\vspace{-3mm}
	\caption{Some examples of the automated preference construction process.}
	\label{fig:example_stage}
\end{figure*}

As illustrated in Figure~\ref{fig:example_stage}, we analyze how each stage of our automated preference construction contributes to the construction of high-quality preference pairs.   
The revision stage improves the sampled responses by correcting factual errors and expanding the depth of the responses. Through critical feedback generation and refinement, potential flaws have the opportunity to be corrected and the responses are enriched with more valuable information.  
The verification stage then quantifies instruction fulfillment capacity of the responses.
Responses which pass the verification are identified as preferred due to their stronger intrinsic alignment with instruction objectives.

\begin{figure*}[t!]
	\centering
	\vspace{-3mm}
	\includegraphics[width=0.94\textwidth]{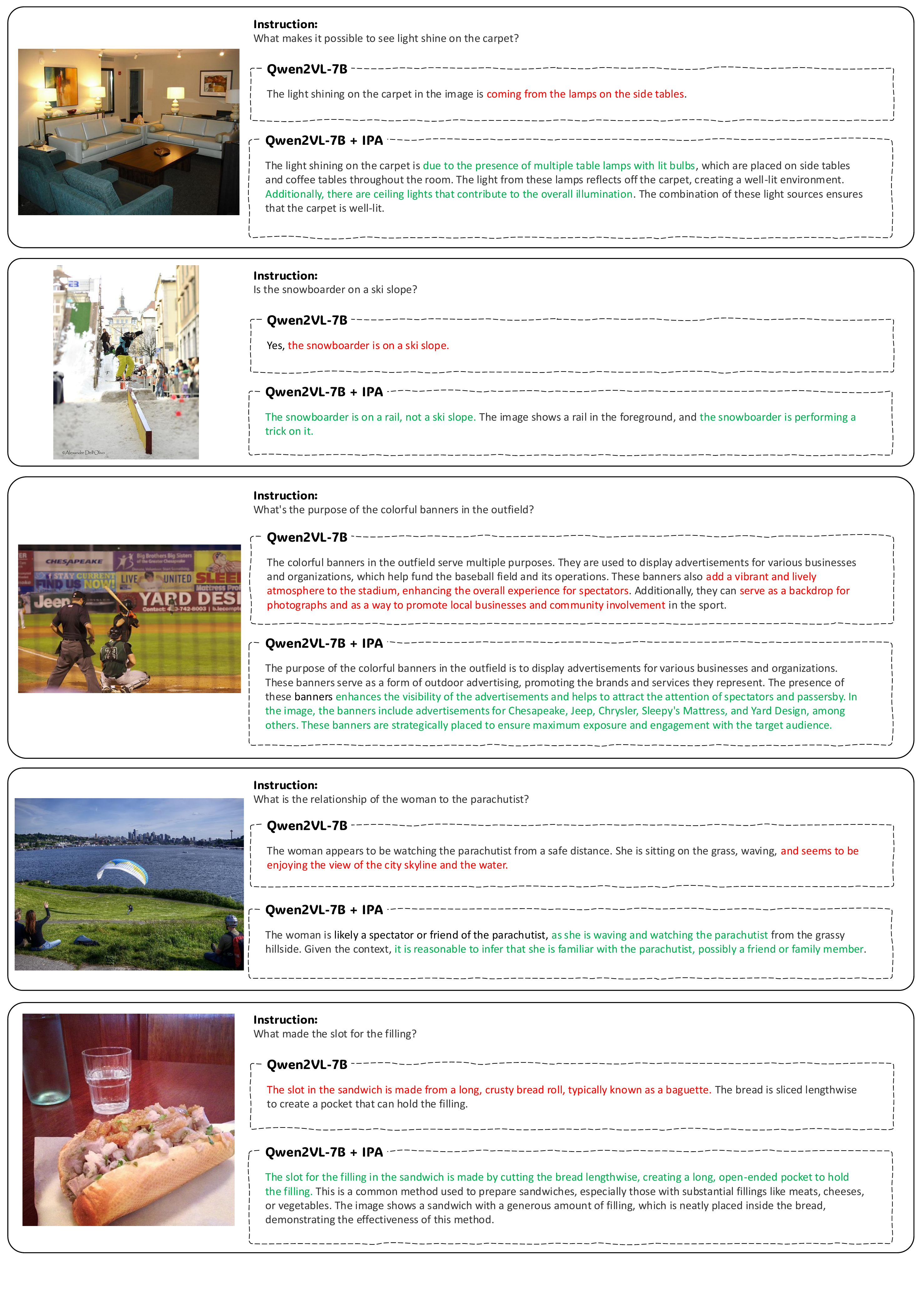}
	\vspace{-6mm}
	\caption{Some examples of the comparison between the baseline model the the model aligned with our approach.}
	\label{fig:example_out}
\end{figure*}

As demonstrated in Figure~\ref{fig:example_out}, we present comparative examples to illustrate the improvements achieved by our approach over the baseline model. 
The visual comparisons reveal that IPA-aligned responses exhibit enhanced capability in addressing the intrinsic demands of instructions.
These examples validate that our approach enables the model to better interpret and fulfill the requirements embedded in diverse instructions, moving  toward more comprehensive understanding capability.

\section{More Experiments}

\subsection{Experiments with More Advanced Models}

We also conduct experiments using larger and more advanced models, InternVL3-14B and Qwen2.5VL-72B, where the same dataset as in the main manuscript is used for training these models. As shown in Table~\ref{tab:table_large_models}, despite the preference dataset being constructed with a relatively weaker and smaller model (Qwen2VL-7B), it brings consistent gains to significantly larger models, demonstrating strong generalizability across model scales and architectures. While the improvements are more moderate due to the already strong baselines, our method still yields meaningful enhancements in overall comprehension ability. We believe that using stronger models for data collection can yield further improvements.

\begin{table*}[h]
\centering
\caption{
    {Prompt $I^{fb}$ for feedback generation.}
}
\begin{minipage}{0.99\linewidth}\vspace{0mm}
\begin{tcolorbox} [colback=white]
    \small

You are a evaluation model tasked with providing a detailed critical analysis of the given instruction-response pair. \\
Please evaluate the response based on the criteria below:\\
\\
\emph{Accuracy of Image Content Description} \\
- Assess whether all descriptions of the image content in the response are accurate. \\
- Identify any instances of hallucinations, where the response describes elements inconsistent with the image.\\
- If any inaccuracies or hallucinations are found, provide a corrected description.\\
    
\emph{Correctness of Knowledge}\\
- Verify the factual accuracy of any commonsense statements or knowledge presented in the response.\\
- If any incorrect knowledge or misconceptions are identified, provide the correct information.\\
    
\emph{Validity of Reasoning and Inferences}\\
- Evaluate the logical consistency of the reasoning processes and inferences made in the response.\\
- Ensure that conclusions are appropriately derived from correct premises.\\
- If any flawed reasoning or incorrect inferences are detected, provide a revised reasoning or inference.\\
    
\emph{Verifiability and Expression of Uncertainty}\\
- Determine if the response includes statements that are difficult to verify or falsify.\\
- Check whether the response clearly indicates uncertainty for such statements when appropriate.\\
- If unverifiable statements are present without indicated uncertainty, suggest how to express the uncertainty clearly.\\
    
\emph{Adherence to Instructions}\\
- Note any deviations from the instructions or omissions of important requirements.\\
- If the response deviates from the instructions or omits key elements, provide a corrected version that fully adheres to the instructions.\\

You should offer feedback detailing all identified issues, followed by a comprehensive evaluation and actionable steps for improvement with direct corrections.\\
\noindent\makebox[\linewidth]{\tikz[baseline]{\draw[dashed] (0,0) -- (\linewidth,0);}}\vspace{2mm}

\textbf{Output Format:}\\
\#\#\# Accuracy of Image Content Description \\
...\\

\#\#\# Correctness of Knowledge \\
...\\

\#\#\# Validity of Reasoning and Inferences\\
...\\

\#\#\# Verifiability and Expression of Uncertainty\\
...\\

\#\#\# Adherence to Instructions\\
...\\

\#\#\# Overall Assessment\\
...\\

\#\#\# Guidance for Improvement\\
...\\
\noindent\makebox[\linewidth]{\tikz[baseline]{\draw[dashed] (0,0) -- (\linewidth,0);}}\vspace{2mm}
\textbf{Input:} \\
\textless Instruction\textgreater \texttt{\{instruction\}}\textless /Instruction\textgreater\\

\textless Response\textgreater \texttt{\{response\}}\textless /Response\textgreater\\

\end{tcolorbox}
\label{tab:prompt_feedback}
\end{minipage}
\end{table*}

\begin{table*}[h]
\vspace{-2mm}
\centering
\caption{
    {Prompt $I^{rev}$ for response refinement.}
}
\begin{minipage}{0.99\linewidth}\vspace{0mm}
\begin{tcolorbox} [colback=white]
    \small

Given an instruction-response pair related to an image, an assistant has provided feedback identifying issues in the response. This feedback enables the generation of an improved response.\\

So your task is to derive a revised response by: \\
$ \text{Response} + \text{Feedback} \rightarrow \text{Revised Response}$ \\

\noindent\makebox[\linewidth]{\tikz[baseline]{\draw[dashed] (0,0) -- (\linewidth,0);}}\vspace{2mm}

\textbf{Output Format:}\\
\#\#\# Revised Response\\
...

\noindent\makebox[\linewidth]{\tikz[baseline]{\draw[dashed] (0,0) -- (\linewidth,0);}}\vspace{2mm}
\textbf{Input:} \\
\textless Instruction\textgreater \texttt{\{instruction\}}\textless /Instruction\textgreater\\

\textless Response\textgreater \texttt{\{response\}}\textless /Response\textgreater\\

\textless Feedback\textgreater \texttt{\{feedback\}}\textless /Feedback\textgreater\\

\end{tcolorbox}
\label{tab:prompt_refinement}
\end{minipage}
\end{table*}

\begin{table*}[h]
\centering
\caption{
    {Prompt $I^{ver}$ for instruction-oriented verification.}
}
\begin{minipage}{0.99\linewidth}\vspace{0mm}
\begin{tcolorbox} [colback=white]
    \small

You are an evaluation model.
Given an instruction-response pair related to an image, your task is to make evaluation by assessing the alignment between the response and a provided reference response.\\

You can assess the alignment by following the criteria below:\\
- Can the reference answer be inferred from the response, i.e., Response $\implies$ Reference?\\
- Are there critical details from the reference missing in the response? \\
- Are there contradictions between the response and the reference? \\
- Does the response adhere to the instruction? \\

\emph{Note:} \\
- It is permissible for the response to contain additional information if it is related to the instruction and not contradictory to the reference.\\
- If this is a multiple-choice/yes-no question, does the response clearly provide the answer? \\

You should indicate whether the response passes the verification (``Pass'' or ``Fail'') concerning all the criteria.

\noindent\makebox[\linewidth]{\tikz[baseline]{\draw[dashed] (0,0) -- (\linewidth,0);}}\vspace{2mm}
\textbf{Output Format:}\\
\#\#\# Assessment\\
...\\

\#\#\# Verification\\
(Pass or Fail)

\noindent\makebox[\linewidth]{\tikz[baseline]{\draw[dashed] (0,0) -- (\linewidth,0);}}\vspace{2mm}
\textbf{Input:} \\
\textless Instruction\textgreater \texttt{\{instruction\}}\textless /Instruction\textgreater\\

\textless Response\textgreater \texttt{\{response\}}\textless /Response\textgreater\\

\textless Reference\textgreater \texttt{\{reference response\}}\textless /Reference\textgreater\\

\end{tcolorbox}
\label{tab:prompt_verification}
\end{minipage}
\end{table*}

\end{document}